\documentclass[letterpaper,10 pt,conference]{ieeeconf}
\IEEEoverridecommandlockouts

\usepackage{graphicx} 
\usepackage{color}
\usepackage{amsmath} 
\usepackage{amssymb}  
\usepackage{enumerate}
\usepackage{cite}
\usepackage{subfigure}
\usepackage[ruled,linesnumbered]{algorithm2e}

\newcommand{\ubf}{\mathbf{u}}\newcommand{\xbf}{\mathbf{x}}

\begin{document}
%
\title{\Large \bf Multi-vehicle Flocking Control with Deep Deterministic Policy Gradient  Method}
%
%
%
\author{Yang Lyu$^{1}$, Quan Pan$^{1}$, Jinwen Hu$^{1}$, Chunhui Zhao$^{1}$ and Shuai Liu$^{2}$
\thanks{This work was supported by the National Natural Science Foundation of China under Grant 61603303, 61473230, the Natural Science Foundation of Shaanxi Province under Grant 2017JQ6005, 2017JM6027, the China Postdoctoral Science Foundation under Grant 2017M610650 and the Fundamental Research Funds for the Central Universities under Grant 3102017JQ02011.} 
\thanks{$^{1}$Yang Lyu, Jinwen Hu,  Chunhui Zhao and Quan Pan are with the Key Laboratory of Information Fusion Ministry of Education, School of Automation, Northwestern Polytechnical University, Xi'an, Shaanxi, 710072, China.
	Email: {\tt\small lincoln1587@mail.nwpu.edu.cn, hujinwen@nwpu.edu.cn,zhaochunhui@nwpu.edu.cn, quanpan@nwpu.edu.cn}. }
\thanks{$^{2}$Shuai Liu is with the School of Control Science and Engineering, Shandong University, Jinan, Shandong, 250061 China. Email:{\tt \small liushuai@sdu.edu.cn}}
}

\maketitle
\begin{abstract}
	Flocking control has been studied extensively along with the wide application  of multi-vehicle systems. In this paper the Multi-vehicles System (MVS) flocking control with collision avoidance and communication preserving is considered based on the deep reinforcement learning framework. Specifically the deep deterministic policy gradient (DDPG) with centralized training and distributed execution process is implemented to obtain the flocking control policy. First, to avoid the dynamically changed observation of state, a three layers tensor based representation of the observation is used so that the state remains constant although the observation dimension is changing. A reward function is designed to guide the way-points tracking, collision avoidance and communication preserving. The reward function is augmented by introducing the local reward function of neighbors. Finally, a centralized training process which trains the shared policy based on common training set among all agents. The proposed method is tested under simulated scenarios with different setup.
\end{abstract}
\section{Introduction}
Multi-Vehicle System  (MVS) raises tremendous research interests in recent years \cite{4140748}. 
Comparing to a single vehicle system, the MVS usually has higher efficiency and operational capability in accomplishing complex tasks such as transportation \cite{alonso2017multi}, search and rescue \cite{liu2016multirobot}, mapping \cite{liu2016towards}. However, the MVS applications also require more sophisticated interaction between vehicles and environemnt, which includes high-level cooperation, competition and behavior control strategies and will significantly increase the system complexity. 
Some of the MVS collective behaviors, such as formation \cite{beard2001coordination} and flocking \cite{olfati2006flocking} are extensively studied in area of control recently with both theoretical analysis and experimental results. However the controller design and analysis method may be unable to deal with the large scale MVS with paralleling multiple purpose interaction, such as way-point tracking, neighboring cooperation and competition, communication preserving and so on.

Recent development of the reinforcement learning (RL) \cite{Sutton2005Reinforcement} has provided an alternative way to deal with the vehicle control problem and shows its potential on applications that involve interaction between multiple vehicles, such as the collision avoidance \cite{chen2017decentralized}, communication \cite{foerster2016learning}, and environment exploration \cite{pham2018cooperative}. Inspired by the works above, we aim to develop an alternative flocking control method by implementing the RL method. 

Some RL based method has already been proposed to deal with the flocking control with collision avoidance under MVS setup. a
hybrid predator/intruder avoidance method for robot flocking combined the reinforcement learning based decision process
with a low level flocking controller has been proposed in \cite{La2015Multirobot}. A complete reinforcement learning based approach for UAV flocking is proposed in \cite{Hung2016A} where the collision avoidance is incorporated into the reward function of the Q-learning scheme. The most similar works related to our topic is proposed in \cite{Long2016Deep} and \cite{chen2017decentralized}.  In \cite{Long2016Deep}, a learning framework imitating the ORCA \cite{Berg2008Reciprocal} is proposed by designing a neural network. However the work is based on the training set from the well validated algorithm which may limit its application to more generalized environment. A deep reinforcement learning method for collision avoidance is proposed in \cite{chen2017decentralized}. Based on the learned policy with designed reward function, the method is validated with improved performance to the ORCA method. 

As the fact that the policy of each agent is dynamically changed in every training loop, which results in un-stationary environment for each agent, the classical Q-learning method is inapplicable. To deal with the policy learning in un-stationary environment with large scale multi-agent system, in this paper we adopt the deep deterministic policy gradient (DDPG) method similar to \cite{DBLP:journals/corr/LoweWTHAM17} with centralized training process and distributed execution process. First,  to avoid the changing size of observation space from different vehicles, a three-layer-tensor with constant size are implemented to represent the observation of neighbors, obstacles and way-points. Then the reinforcement learning function with collision avoidance, way-points tracking and communication preserving are designed. To further take into consideration of the state of neighbors, the reward function are augmented with the reward of neighbors  with a discount factor. Finally the DDPG is trained and the replay buffer is filled with all vehicle’s state transition which means the training process is centralized and the policy is shared among all vehicles.

The remainder of this paper is organized as follows. In Section II the basis idea of deep reinforcement learning method is described. The system modeling and problem description is described in Section III.  In Section IV our reinforcement learning based method is proposed and Section V validates the proposed method based on experiments. Section VI concludes this paper.

%
\section{Deep Reinforcement Learning}
\subsection{Deep Q-Learning}
In reinforcement learning, an agent receives the current state of the environment $s^k\in \mathcal{S}$ and selects an action $a^k\in \mathcal{A}$ based on this state according to a stochastic policy $\pi\left(a^k|s^k\right)$ or a deterministic policy $a^k = \mu\left(s^k\right)$, then the agent receives a reward $r^k = R\left(s^k,a^k\right) \in \mathcal{R}$ and arrives at a new state $s^{k+1} = s'$. 
The transition dynamics are in general take as Markovian, with the transition probability
\begin{equation}
p\left(s'|s,a\right) = {\rm{Pr}}\{s^{k+1}=s'|s^k= s,a = a^k \}.
\end{equation} 
Obviously the reinforcement learning problem can be treated as a Markov Decision Process (MDP) with the tuple $\langle\mathcal{S},\mathcal{A},P, R,\gamma\rangle$. $\gamma$ is a discount parameter.
The core objective is to find a policy which maximizes the cumulative long term discount gain 
\begin{equation}
\label{obje}
G_k = \sum\limits_{t=0}^{\infty}\gamma^k r^{k+t}, \quad \gamma \in \left[0,1\right).
\end{equation}
Specifically, in Q-learning, the value of $G$ by taking a certain action $a$ from state $s$ is called Q-function, which is progressively updated with 
\begin{equation}
Q\left(s,a\right) = Q\left(s,a\right) + \alpha \left(r+ \gamma \max\limits_{a'}Q\left(s',a'\right) - Q\left(s,a\right) \right).
\end{equation}
The traditional RL algorithm are typically limited to discrete, low dimensional domains and poorly suited to multi-vehicle environment with continuous state/action and observation. 

Recent advances in deep reinforcement learning  \cite{Mnih2015Human} have demonstrated human-level performance in complex and high-dimensional spaces. As an extension of Q-learning on high-dimension application, the deep Q-learning method uses a deep neural network with parameters $\theta^Q$ to approximate the Q-function on continuous state and discrete action space.
By defining the lost function 
\begin{equation}
\label{lost}
L\left(s,a|\theta^Q\right) = \left(y-Q\left(s,a|\theta^Q\right)\right)^2
\end{equation}
with the target values
\begin{equation}
y = r+ \gamma \max\limits_{a'}Q\left(s',a'\right)
\end{equation}
The parameter $\theta^Q$ is updated using back-propagation
\begin{equation}
\theta^Q = \theta^Q + \alpha \nabla_{\theta^Q}L\left(s,a|\theta^Q\right)
\end{equation}
One of the update rule is presented as the ADAM\cite{Kingma2014Adam}.

\subsection{Deep deterministic policy gradient}
The deep Q-learning method above may be difficult to extended to the multi-vehicle environment as the local policy is changing based on local training process, the learning process based on global training set maybe unstable, therfore a global $Q$ function is not feasible. To overcome the drawback, the deep deterministic policy gradient (DDPG) method can be implemented by introducing extra target network for the $Q\left(s,a|\theta^Q\right)$ function and a deterministic actor function $\mu(s,|\theta^{\mu})$, as $Q'$ and $\mu'$ and a replay buffer. The DDPG updates as follows.

Define the target value as
\begin{equation}
y = r+\gamma Q'(s',u'(s'|\theta^{\mu'})|\theta^{Q'})
\end{equation}
Update the critic by minimizing the lost function
\begin{equation}
\label{begin}
L\left(s,a|\theta^Q\right) = \left(y-Q\left(s,a|\theta^Q\right)\right)^2
\end{equation}
Given the objective function similar to (\ref{obje}), the gradient can be calculated as
\begin{equation}
\nabla_{\theta_{\mu}}G(\theta^\mu) = \nabla_{\theta^\mu_k}\mu(s|\theta^\mu_k)\nabla_aQ(s,a|\theta^Q_k)|_{a = \mu(s|\theta^\mu_k)}
\end{equation}
Then the parameters $\theta^Q$ and $\theta^\mu$ can be updated respectively as 
\begin{align}
\theta^Q &= \theta^Q + \alpha\nabla_{\theta^Q}L(\theta^Q)\\
\theta^\mu &= \theta^\mu + \alpha \nabla_{\theta^\mu}G(\theta^\mu)
\end{align}
The parameters of target network is updated after a sequence of $S$ critic and actor network update as
\begin{align}
\theta^{Q'}\leftarrow\tau \theta^{Q} + (1-\tau) \theta^{Q'} \\
\label{end}\theta^{\mu'}\leftarrow\tau \theta^{\mu} + (1-\tau) \theta^{\mu'} 
\end{align} 
where $\alpha $ and $\tau$ is to determine the update rate of the actor-critic network and target network respectively.

For our multi-vehicle application, we implement the centralized training and decentralized execution process. In the centralized training process, each critic network is augmented with policies of its neighbors by including the reward of neighbors and trained based on shared training set.   For the decentralized execution process, decision is made based on observation of neighbor states.
\section{System Modeling and Problem}
In this section the MVS is modeled using unicycle model and proximity network and the multi-vehicle control problem involving way-points tracking, collision avoidance and communication preserving is described.
\subsection{System Modeling}
\subsubsection{Dynamic Model}
In this paper we consider a set of $n$ homogeneous mobile vehicles, denote as $\mathcal{V}$, are operating in 2D space with unicycle model. The discretized dynamic model for each vehicle $i$ is described as 
\begin{equation}\label{dynamic}\left\{
\begin{split}
x_i^{k + 1} &=x_i^{k} + v_i^{k}\Delta t \cos\left(\Theta^{k}\right)\\
y_i^{k + 1} &= y_i^{k} + v_i^{k} \Delta t\sin\left(\Theta^{k}\right)\\
\Theta_i^{k + 1} &= \Theta_i^{k} + \omega_i^{k}\Delta t
\end{split}\right.
\end{equation}
where $x,y,\Theta$ denote  the position and the heading angle respectively in 2D space. $v_i$ and $w_i$ are respectively the linear velocity and angular velocity. We define the position of vehicle $i$ at time instance $k$ as $\xbf_i^k = \left[x_i^k,y_i^k\right]^\top$ and control input as $\ubf_i^k = \left[v_i^k, w_i^k\right]^\top$. $\Delta t$ denotes the sampling period.
\subsubsection{Proximity Network}
In this paper, an undirected proximity graph $\mathcal{G}_c^k=\left(\mathcal{V},\mathcal{E}_c^k\right)$ is used to represent the communication topology of the multi-vehicle system at each time instance $k$, where $\mathcal{V}$ and $\mathcal{E}^k \in \mathcal{V} \times \mathcal{V} $ are, respectively, the set of vertices that stands for the local vehicles and the edge set that stands for the communication links. In proximity network, the edge set $\mathcal{E}^k$ is defined according to the spatial distance between vehicles, namely $d_{ij}^k =\|\xbf^k_i - \xbf^k_j\|$,  as 
\begin{equation}
\mathcal{E}_c^k=\{\left(i,j\right)\left.\right|d_{ij}^k<r_n, i,j\in \mathcal{V}, i\ne j\},
\label{graph}
\end{equation}
where $r_n$ is the proximity network threshold. The neighborhood set of vehicle $i$ is defined as ${N}_i^k \triangleq \{j|(i,j)\in \mathcal{E}_c^k\}$. 

Besides the graph $\mathcal{G}_c^k$, an additional directed graph $\mathcal{G}_o^k(\mathcal{V}\cup \mathcal{O}, \mathcal{E}_o^k)$ is also implemented to represent the MVS obstacle  detection  status, where $\mathcal{O}$ is the set of obstacle and the edge set $\mathcal{E}_o^k\in \mathcal{V}\times \mathcal{O}$ denote the pairwise detection between vehicles and obstacles. Similar, the existence of edge depends on the detection range of each vehicle $r_o$, which is defined similar as 
\begin{equation}
\mathcal{E}_o^k=\{\left(i,o\right)\left.\right|d_{io}^k<r_o, , i\in \mathcal{V}, o\in \mathcal{O}\},
\label{graph_o}
\end{equation}
where the spatial distance between vehicles and obstacle is $d_{io}^k =\|\xbf_i^k-\xbf_o^k\|$. 
The obstacles that within vehicle $i$'s sensing range is defined as $C_i^k\triangleq\{{o|(i,o)\in \mathcal{E}_o^k}\}$.
\subsection{Problem description}
The objective is to develop distributed local controllers with way-point tracking, collision avoidance and network preserving.
\begin{itemize}
	\item {\it Way-point tracking}: Given the discretized way-points $\xbf_r^k$, the control object is to minimize the weighted tracking error norm $e_{t} = \|\xbf_i - \xbf_r\|$.
	\item {\it Collision avoidance}: Given a predefined minimum separation distance $r_n'$ and $r_o'$, the relationship between vehicle $i$ and vehicle $j$ or vehicle $i$ and obstacle $o$ should satsifies
	\[d_{ij}^k\ge r_n', i,j\in \mathcal{V},\]
	\[d_{io}\ge r_o', i\in \mathcal{V}, o\in C_i^k.\]
	\item{\it Communication Preserving}: Given a maximum communication range $r_n$ between two vehicles, the objective is to keep the connectivity of the graph by driving the vehicles stay within the sensing range of each other, as 
	\[d^k_{ij}\le r_n, i\in \mathcal{V}, j\in N^k_i.\]
\end{itemize}
Similar problem setup is extensively studied in \cite{olfati2006flocking} based on controller design and stability analysis. In this paper we would like to solve the problem in an alternative way without explicit analytical process. Instead, we formulate the above three objectives and the dynamic model using the reinforcement learning scheme. Based on the centralized training process and distributed execution process, similar behavior to the flocking \cite{olfati2006flocking} can be achieved.
\section{Flocking control with DRL}
In this section, we introduce the key ingredients of our reinforcement learning based flocking control framework. Specifically we begin with representing the observed state as three layers tensor, then the reward function is described with regard to our flocking control objective. The DDPG based flocking control is described in the end.
\subsection{Observation Representation}
For the MVS, the interaction between agent $i$ and the environment contains three aspects, namely, the status of its cooperative neighbor $j\in N_i$, the obstacle status within its sensing range $o\in C_i$, the common way-points $W$.
In order to model the status in a continuous space at the same time remain the observation space invariant, we represent the three types of observation mentioned above as three channel of image-like tensor within different scale. 
\begin{figure} 
	\centering 
	\subfigure[Neighborhood status]{ 
		\label{fig:subfig:a} 
		\includegraphics[width=1.5in]{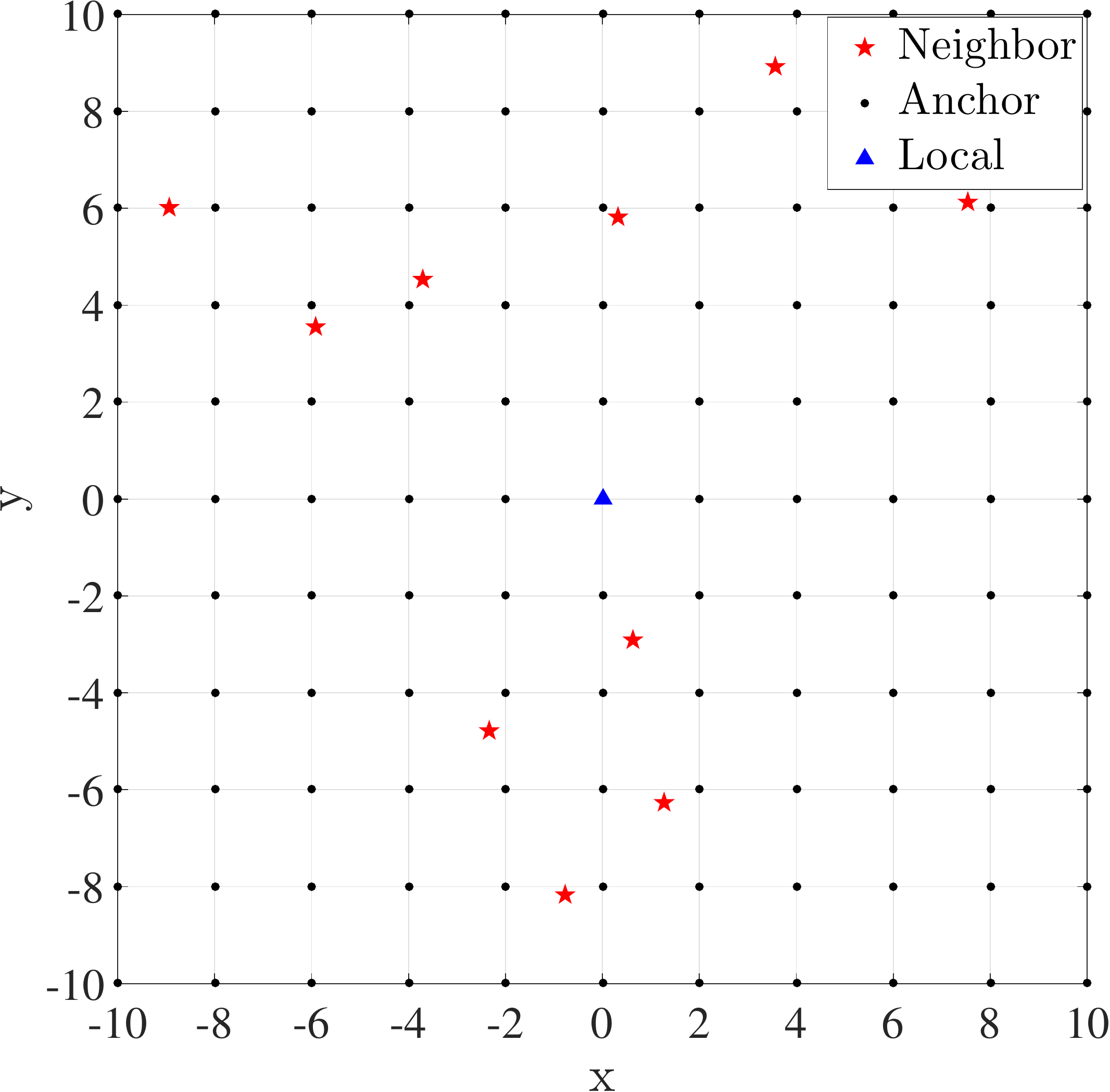} 
	} 
	\subfigure[Observation]{ 
		\label{fig:subfig:b} 
		\includegraphics[width=1.5in]{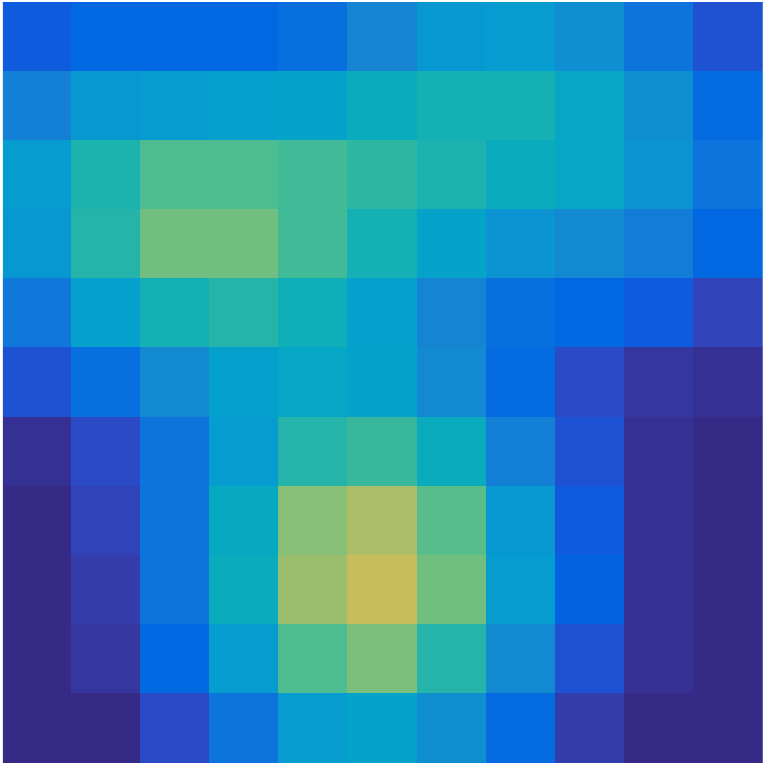} 
	} 
	\caption{The neighborhood status in  the coordinate of the local vehicle and the corresponding observation representation} 
	
	\label{fig:subfig} 
\end{figure}

\paragraph{Neighbor channel}: In this channel, the location of neighborhood agents in the local frame of the vehicle $i$， defined as $\xbf_{ij}\in \mathbb{R}^2, j\in \mathcal{N}_i$, are projected to a 2D matrix $I_N$. The transformation is as follows. First $ l\times l$ virtual points, namely the anchors, are equally distributed within the area $\left[\pm r_c, \pm r_c\right]$, then each anchor is able to measure the neighbor intensity around it based on a Gaussian radial function as 
\begin{equation}
\small	
\varphi\left(\xbf_{ij},\pmb\mu_{m,n}\right) = \exp \left(\left(\xbf_{ij}-\pmb \mu_{m,n}\right)^\top \Sigma^{-1}\left(\xbf_{ij}-\pmb \mu_{m,n}\right)\right),
\label{gaussian}
\end{equation}
where $\xbf_{io}$ denote the neighbor $j$'s location in the local frame of agent $i$, and is calculated as
$\xbf_{ij} = R(\Theta_i)(\xbf_j -\xbf_i)$.
$\pmb \mu_{m,n}$ denotes the location of anchor $(m,n)$ in the local frame of agent $i$, $\Sigma$ represent the anchor's sensitivity to each vehicle's radian.  The measurement value of anchor is the summation of all neighbors' radiation on anchor $i$, that is  
\begin{equation}
I_N\left(m,n\right) = \sum\limits_{j\in N_i}\varphi\left(\xbf_{ij},\pmb \mu_{m,n}\right)
\end{equation}
An example of the data representation is as figure \ref{fig:subfig} and there are $10$ vehicles within the sensing range of vehicle $i$'s sensing range, denote as the neighbors ${N}_i$, as Figure \ref{fig:subfig:a}. By implementing a $11\times 11$ anchor grid, the neighbor observation $I_N$ based on above definition is represented in the color domain as Figure \ref{fig:subfig:b}.

\paragraph{Obstacle channel} In  this channel, The location of the  obstacles within the agent's sensing range $r_o$ in the local frame are compressed into a similar channel within the area $\left[\pm r_o,\pm r_o\right]$ as the neighbor status.
\begin{equation}
I_O\left(m,n\right) = \sum\limits_{o\in C_i}\phi\left(\xbf_{io}, \pmb \nu_{m,n}\right),
\end{equation}
where $\phi\left(\cdot\right)$ is a similar Gaussian radial function as $\phi$ in Eq.~\ref{gaussian} and $\pmb \nu_{m,n}$ is the  anchor $\left(m,n\right)$'s position in the local frame of agent $i$, $\xbf_{io}$ denotes the obstacle $o$'s position in vehicle $i$'s local frame, as $\xbf_{io} = R_{i}(\theta_i)(\xbf_o -\xbf_i)$.
\paragraph{Goal channel} In this channel, the next waypoints are projected onto a similar channel with anchors equally positioned in the region $\left[\pm r_g,\pm r_g\right]$, and the observation $I_W$ is calculated as
\begin{equation}
I_W\left(m,n\right) = \psi\left(\xbf_{ig},\pmb\omega_{m,n}\right).
\end{equation}
\subsection{Reward function}
The reward function is concerning three aspect of the our objective, namely connectivity preserving, obstacle avoid and way-points tracking which are detailed as follows:
\begin{itemize}
	\item{\bf{Connectivity Preserving}}:
	This function is to maintain the distance between each vehicle and its neighbor within the maximum communication range $r_n$ at the same time keep a minimum separation distance $r_n'$. Consequently the pairwise reward function for vehicle $i$ and its neighbor $j\in N_i$ can be defined as
	\begin{equation}
	\delta^N_{ij} = 
	\begin{cases}
	1 & r_n' \le \|x_j^i\|\le r_n, \\
	-1 & \|\xbf_{ij}\| < r_n',\\
	0 & \rm{others}.
	\end{cases}
	\end{equation}
	\item{\bf{Obstacle Avoidance}} This function is to avoid collision with obstacle, that is, to keep a minimum separation distance $r_o'$ between vehicle $i$ and the obstacle $o$ within its sensing range, $o\in C_i$. The pairwise reward function is defined as
	\begin{equation}
	\delta^O_{io} = 
	\begin{cases}
	-1 & \|\xbf_{io}\| < r_o',\\
	0 & \rm{others}.
	\end{cases}
	\end{equation}
	\item{\bf{Way-points Tracking}}: The agents should follow the predefined mission way-points. The reward is defined based on a normalized distance of local vehicle $i$ to the way-point,
	\begin{equation}
	\delta^W = -{\epsilon\|\xbf_{ig}\|},
	\end{equation}
	where $\epsilon$ is a normalized factor of the distance between target way-point and position of an agent. 
\end{itemize}
Finally the reward function to evaluate the behavior of the agent $i$ is composed as 
\begin{equation}
r_i = \sum\limits_{j\in N_i}\delta_{ij}^N + \sum\limits_{k\in C_i}\delta_{ik}^O + \delta^W +\beta\|\ubf_i\|^2
\label{rew_i}
\end{equation}
The last term is a punish term to enforce smooth action trajectory and economic maneuver with a negative weight factor $\beta$.
In the MVS operation environment, we defined a inclusive reward function $r_{\{i\}}$ as a combination of the reward function (\ref{rew_i}) and the discounted reward of neighbors $j\in N_i$ as 
\begin{equation}
r_{\{i\}} = \lambda r_i + (1-\lambda) \frac{1}{|N_i|}\sum_{j\in N_i}r_j,
\end{equation}
where $\lambda$ is a weight factor denotes how much portion of the interest of neighbors are considered.
\subsection{DDPG network}
According to the DDPG framework proposed in \cite{DBLP:journals/corr/LoweWTHAM17},  we would like to derive a policy learning method based on following setup: (1) the input of the learning policy is based on local observation on neighbors, obstacles as well as way-points and (2) the experiences are collected from all vehicles and to train a shared policy, which make the framework involves a centralized training process and a distributed execution process. According to the Algorthm 1, the reward during each action is defined as the vehicle $i$ as well as its neighbors. The replay buffer is filled with all vehicles' state transition which means the training process is centralized and the policy is shared among all vehicles.

\begin{figure}
	\centering
	\includegraphics[width=1\linewidth]{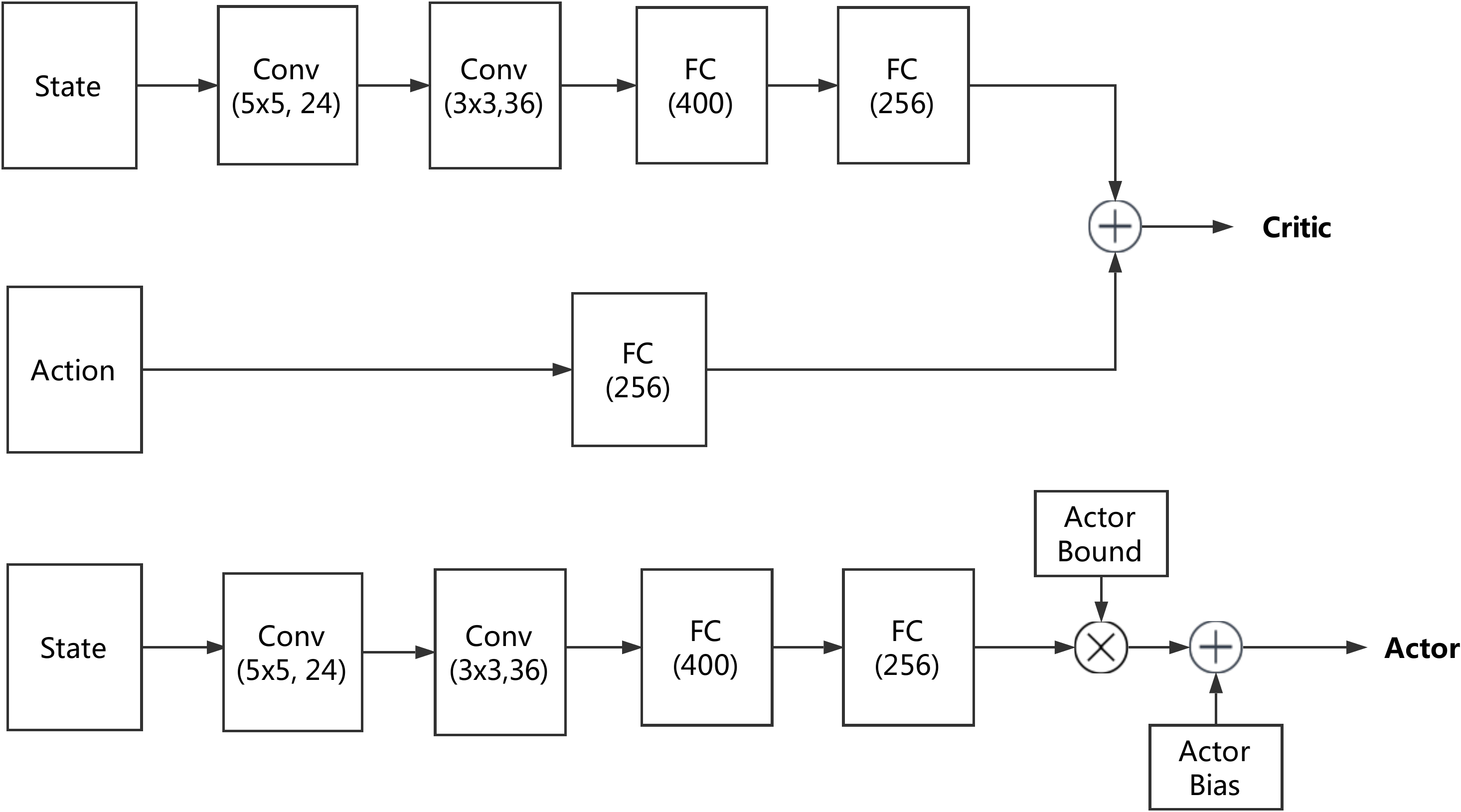}
	\caption{The DDPG netowrk struchture used for our MVS control, consist of a critic network and a actor network.}
	\label{fig:Critic_Actor-crop}
\end{figure}

The critic network and actor network is represented as Figure \ref{fig:Critic_Actor-crop}. The critic network contains four hidden  layers neural network for the state and one hidden layer for the action to approximate the action-state function $Q(s,a|\theta^Q)$, the actor network contains similar four-layer network appended with the action bias and action bound to approximate the actor network $\mu(s|\theta^{\mu})$.
Specifically, the two convolution layers are implemented as pre-process of the observation and each convolution layers contains multiple convolution kernels and ReLu layers. The overall algorithm of the described DDPG based flocking control framework is presented as Algorithm 1.
\begin{algorithm}
	\caption{Deep Deterministic Policy Gradient based Flocking}
	Initialize critic network $Q\left(a,s|\theta^Q\right)$ and actor $\mu^{\theta}\left(s|\theta^\mu\right)$ with parameters $\theta^Q$ and $\theta^\mu$.
	Initialize target network $Q'$ and actor $\mu'$ with parameters $\theta^{Q'}\leftarrow\theta^Q$ and $\theta^{\mu'}\leftarrow\theta^\mu$.
	Initialize the replay buffer $\mathcal{R}$.\\
	\For {\rm{episode =1,M}}{
		initilize the environment setup\\
		\For {${k=1,T}$}
		{\For {$i \in \mathcal{V}$ in parallel}{
				Receive observation $\xbf^k_{ij},\xbf^k_{io},\xbf^k_{ig}$\\
				Represented observation as $s^k=\{I^k_N, I^k_O, I^k_G\}$ \\
				Select action according to $a_k^i = \mu\left(h_t\right)+\varepsilon$\\
				Execute action $a^k_i$, receive reward $r_i$ and reach state$s_i^{k+1}$\\
				Collect reward from $j\in N_i$ and obtain the inclusive reward $r_{\{i\}}$\\
				Store the tuple $\{s_i^k, a_i^k,r_{\{i\}}^k, s_i^{k=1}\}$ into $\mathcal{R}$\\
				Sample mini-batch with length $S$ from $\mathcal{R}$\\
				DDPG update according to (\ref{begin}) to (\ref{end})}
		}
	}
\end{algorithm}
\begin{table}
	\centering
	\caption{The test results of 1000 episodes in Experiment 1}
	\label{my-label}
	\begin{tabular}{|l|l|}\hline
		Reference way-points tracking Error   & 0.08\\ \hline
		Minimum separation distance to obstacle & 0.138 \\\hline
		Minimum separation distance to neighbors & 0.094\\\hline
		Average separation distance to neighbors & 0.185\\\hline 
	\end{tabular}
\end{table}
\begin{table*}[tbh]
	\centering
	\caption{The test results under different scenario setup}
	\label{tab2}
	\begin{tabular}{|c|c|c|c|c|c|}
		\hline
		\multicolumn{2}{|c|}{Setup} & \multicolumn{4}{c|}{The flocking performance}                                                                                                                     \\ \hline
		No. vehicles & No. obstacles & Aver. training time (ms) & W.P. tracking error & Min. sep. dis. (obs./nei.) & Aver. sep. dis. to nei.\\ \hline
		3            & 1            & 122                             & 0.125                         & 0.158/0.102                                         & 0.187                                     \\ \hline
		5            & 1            & 140                             & 0.152                         & 0.152/0.110                                         & 0.174                                     \\ \hline
		5            & 2            & 147                             & 0.147                         & 0.144/0.103                                         & 0.182                                     \\ \hline
	\end{tabular}
\end{table*}
\section{experiment}
In this section, the experiments and results of the proposed flocking control framework are presented. We setup the flocking control environment with different number of cooperative agent  and uncooperative obstacles. The object is to track the desired reference waypoint at the same time maintain a proper distance with its neighbors and avoid a random moving obstacles.
In particular, our DDPG network is designed based on the tensorflow deep learning framework \cite{abadi2016tensorflow} and the scenario is built using the Gym package \cite{brockman2016openai} and the multiagent particle environment (MPE) package\cite{lowe2017multi}. The python implementation of our algorithm is carried out on a laptop with i7-7700HQ CPU and GTX 1050 graphic card.

During the experiments, the vehicles and obstacles move according to the dynamic model (\ref{dynamic}) with sample time $\Delta t = 0.1$. Specifically, the maximum velocity for the vehicles and obstacle are $0.15$ and $0.1$ respectively. The minimum separation distance to neighbors and obstacle are  set as $r_n' = 0.1$ and $r_o' = 0.15$. The distance threshold for the proximity network $\mathcal{G}_c^k$ and $\mathcal{G}_o^k$ are respectively set as $r_n = 0.15$ and $r_o = 0.25$.
Initially the position of vehicles are placed randomly in the 2D plane $\{(x,y)|-1\le x,y\le 1, \}$ with random $v_i^0$ and $\Theta_i^0$. The obstacle is placed in the same area and subject to a random walking process.  The reference waypoint is randomly placed with constant velocity 0.1 towards the original points. 

\begin{figure}
	\centering
	\includegraphics[width=0.9\linewidth]{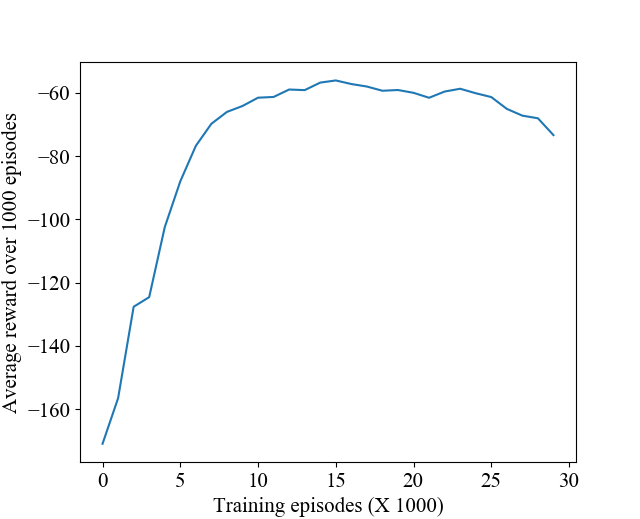}
	\caption{The vehicles' averaged reward over 1000 episodes with total 30000 training episodes}
	\label{fig:training_curve}
\end{figure}

\paragraph{Experiment 1}
In this experiment, the collision avoidance, reference waypoint tracking and obstacle avoidance capabilities based on our proposed method is evaluated. The scenario is set as 3 vehicles and one obstacle. The training set is set as 30000 episodes, and we use 1000 episodes to evaluate the performance. The averaged training time is 122ms per episode. The averaged reward over every 1000 episode is plotted in Figure \ref{fig:training_curve}, which shows a stable flocking control policy is obtained after 10000 episodes. The test results of 1000 episodes are shown in Table \ref{my-label}. Obviously the collision avoidance and way-points tracking is demonstrated.
\begin{figure}
	\centering
	\includegraphics[width=0.75\linewidth]{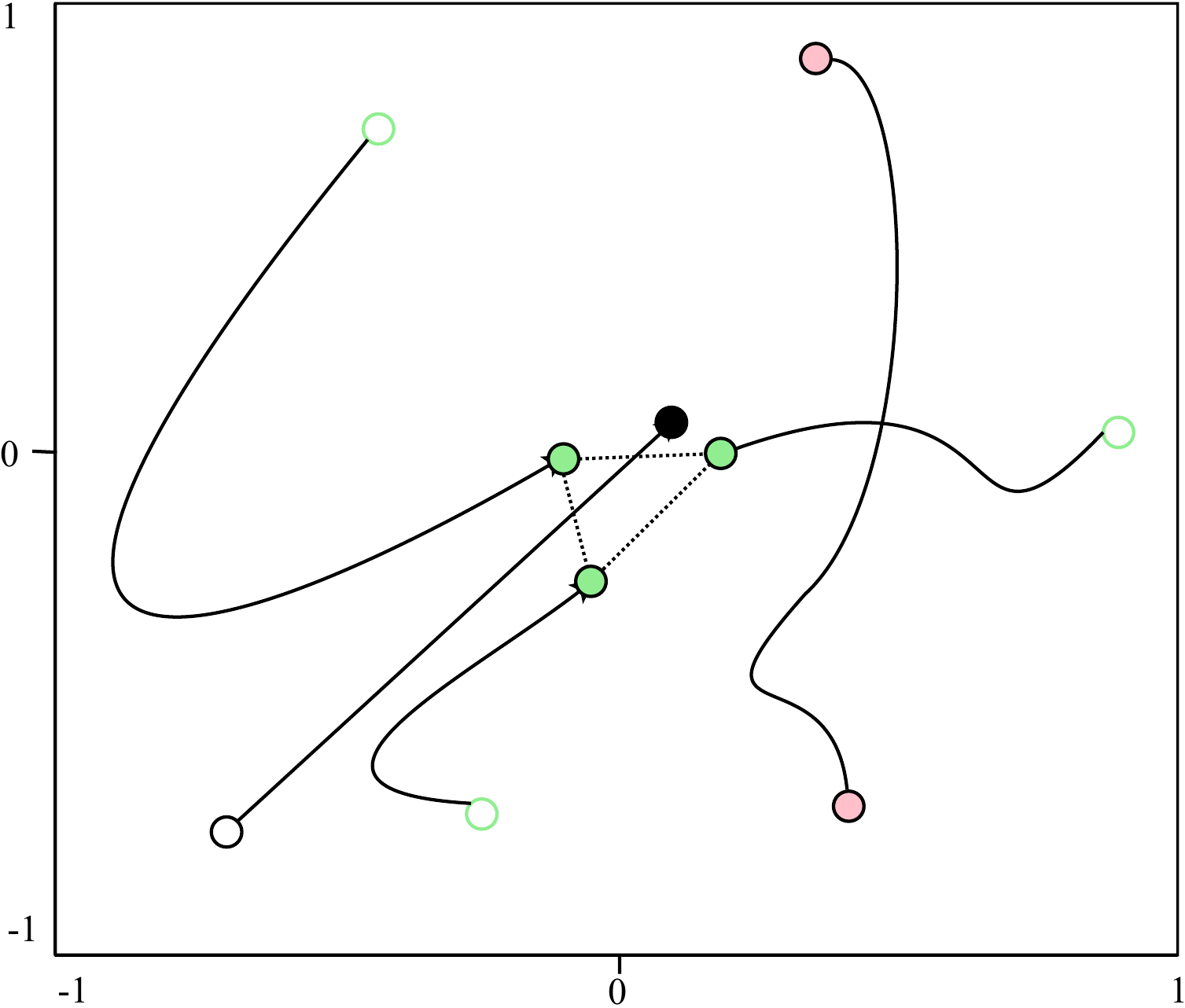}
	\caption{The trajectories of 1 obstacle (pink), 3 vehilces (green) and the reference waypoint (black).}
	\label{fig:traj-crop}
\end{figure}

\paragraph{Experiment 2}
In this experiment, different scenarios are set to evaluate the performance of our method. The results is shown in Table. II. Three different scenarios are defined based on different vehicles and obstacles. Apparently the training time shows only slight increase  as the number of vehicles and obstacles grow, which mainly because that the state space remain constant with the variation of number of vehicles or obstacle based on our observation representation method. The way-point tracking error, collision avoidance and communication preserving are both demonstrated with similar statistic performance.

\section{Conclusion}
In this paper, a reinforcement learning framework for flocking control with collision avoidance and communication preserving are proposed. The main differences of our work lies in two folds, 1) we implemented a observation represented method which transform the state with dynamically changed size into a tensor based state which remain constant, and 2) we design a centralized training framework which uses the augmented reward function and shared policy which is trained based on the common replay buffer filled by all vehicles. The experiment results show that the proposed method is able to demonstrate the flocking control with acceptable performance.
In the further the work will be extended with more experiments with detailed analysis. As one possible directions, the theoretical analysis of the consensus on policy will be carried out.
\section*{Acknowledgment}
The authors would like to thank the National Research Foundation, Keppel Corporation, and National University of
Singapore for supporting this work done in the Keppel-NUS Corporate Laboratory. The conclusions put forward reflect the
views of the authors alone and not necessarily those of the institutions within the Corporate Laboratory. The WBS number
of this project is R-261-507-004-281.
\bibliographystyle{IEEEtran}
\bibliography{my_bib}
\end{document}